\documentclass{article} %
\usepackage{nips13submit_e,times}
\usepackage{hyperref}
\usepackage{url}
\usepackage{verbatim}
\usepackage{graphicx}
\usepackage{caption}
\usepackage{subcaption}

\usepackage{algorithm,algpseudocode}

\usepackage{amsthm}
\usepackage{amsmath}
\usepackage{amsfonts}
\usepackage{amssymb}

\theoremstyle{definition}

\title{Auto-Encoding Variational Bayes}

\author{
Diederik P. Kingma \\
Machine Learning Group \\
Universiteit van Amsterdam \\
\texttt{dpkingma@gmail.com} \\
\And
Max Welling \\
Machine Learning Group \\
Universiteit van Amsterdam \\
\texttt{welling.max@gmail.com} \\
}

\usepackage{amsthm}
\usepackage{amsmath}
\usepackage{amsfonts}
\usepackage{amssymb}

\newcommand{\bb}[1]{\mathbf{#1}}
\newcommand{\bbb}{\bb{b}}
\newcommand{\bx}{\bb{x}}
\newcommand{\bxi}{\bx^{(i)}}
\newcommand{\bxl}{\bx^{(l)}}

\newcommand{\by}{\bb{y}}

\newcommand{\bg}{\bb{g}}
\newcommand{\bh}{\bb{h}}

\newcommand{\bz}{\bb{z}}
\newcommand{\bzi}{\bz^{(i)}}
\newcommand{\bzl}{\bz^{(l)}}
\newcommand{\bzil}{\bz^{(i,l)}}

\newcommand{\bT}{\boldsymbol{\theta}}

\newcommand{\balpha}{\boldsymbol{\alpha}}
\newcommand{\bphi}{\boldsymbol{\phi}}
\newcommand{\beps}{\boldsymbol{\epsilon}}
\newcommand{\bepsl}{\beps^{(l)}}
\newcommand{\bepsil}{\beps^{(i,l)}}
\newcommand{\bzeta}{\boldsymbol{\zeta}}
\newcommand{\bzetal}{\zeta^{(l)}}
\newcommand{\bsigma}{\boldsymbol{\sigma}}
\newcommand{\bmu}{\boldsymbol{\mu}}
\newcommand{\bzero}{\bb{0}}

\newcommand{\bW}{\bb{W}}
\newcommand{\bI}{\bb{I}}

\newcommand{\bX}{\bb{X}}

\newcommand{\btz}{\widetilde{\bz}}
\newcommand{\btT}{\widetilde{\bT}}

\newcommand{\pT}{p_{\bT}}

\newcommand{\pA}{p_{\balpha}}
\newcommand{\qT}{q_{\bT}}
\newcommand{\qPhi}{q_{\bphi}}

\newcommand{\fPhi}{f_{\bphi}}

\newcommand{\gPhi}{g_{\bphi}}
\newcommand{\hPhi}{h_{\bphi}}

\newcommand{\Exp}[2]{\mathbb{E}_{#1}\left[#2\right]}

\newcommand{\eqnr}{\addtocounter{equation}{1}\tag{\theequation}}

\theoremstyle{definition}

\newcommand{\LB}[2]{\mathcal{L}^{#1}(\bT,\bphi; #2)}
\newcommand{\LBT}[2]{\widetilde{\mathcal{L}}^{#1}(\bT,\bphi; #2)}

\usepackage{tikz}
\usetikzlibrary{bayesnet}

\nipsfinalcopy %

\begin{document}

\maketitle

\begin{abstract}
How can we perform efficient inference and learning in directed probabilistic models, in the presence of continuous latent variables with intractable posterior distributions, and large datasets?
We introduce a stochastic variational inference and learning algorithm that scales to large datasets and, under some mild differentiability conditions, even works in the intractable case. Our contributions are two-fold.
First, we show that a reparameterization of the variational lower bound yields a lower bound estimator that can be straightforwardly optimized using standard stochastic gradient methods.
Second, we show that for i.i.d. datasets with continuous latent variables per datapoint, posterior inference can be made especially efficient by fitting an approximate inference model (also called a recognition model) to the intractable posterior using the proposed lower bound estimator. 
Theoretical advantages are reflected in experimental results.
\end{abstract}

\section{Introduction}

How can we perform efficient approximate inference and learning with directed probabilistic models 
whose continuous latent variables and/or parameters have intractable posterior distributions?
The variational Bayesian (VB) approach involves the optimization of an approximation to the intractable posterior. Unfortunately, the common mean-field approach requires analytical solutions of expectations w.r.t. the approximate posterior, which are also intractable in the general case. We show how a reparameterization of the variational lower bound yields a simple differentiable unbiased estimator of the lower bound; this SGVB (Stochastic Gradient Variational Bayes) estimator can be used for efficient approximate posterior inference in almost any model with continuous latent variables and/or parameters, and is straightforward to optimize using standard stochastic gradient ascent techniques.

For the case of an i.i.d. dataset and continuous latent variables per datapoint, we propose the Auto-Encoding VB (AEVB) algorithm. In the AEVB algorithm we make inference and learning especially efficient by using the SGVB estimator to optimize a recognition model that allows us to perform very efficient approximate posterior inference using simple ancestral sampling, which in turn allows us to efficiently learn the model parameters, without the need of expensive iterative inference schemes (such as MCMC) per datapoint. The learned approximate posterior inference model can also be used for a host of tasks such as recognition, denoising, representation and visualization purposes. When a neural network is used for the recognition model, we arrive at the \emph{variational auto-encoder}.

\section{Method}
\label{sec:method}

The strategy in this section can be used to derive a lower bound estimator (a stochastic objective function) for a variety of directed graphical models with continuous latent variables. We will restrict ourselves here to the common case where we have an i.i.d. dataset with latent variables per datapoint, and where we like to perform maximum likelihood (ML) or maximum a posteriori (MAP) inference on the (global) parameters, and variational inference on the latent variables. It is, for example, straightforward to extend this scenario to the case where we also perform variational inference on the global parameters; that algorithm is put in the appendix, but experiments with that case are left to future work. Note that our method can be applied to online, non-stationary settings, e.g. streaming data, but here we assume a fixed dataset for simplicity.

\begin{figure}[t]
\begin{center}
\begin{tikzpicture}[scale=1, transform shape]
\node[obs] (x1) {$\mathbf{x}$};
\node[latent, above=of x1] (z1) {$\mathbf{z}$};
\node[const, left=of z1] (phi1) {$\mathbf{\phi}$};
\node[const, right=of z1] (theta1) {$\mathbf{\theta}$};
\edge [dashed] {phi1} {z1};
\edge {theta1} {z1};
\edge {theta1} {x1};
\draw (x1) edge[out=135,in=225,->,dashed] (z1);
\edge {z1} {x1};
\plate [xscale=1.5] {} {(x1)(z1)} {$N$} ;
\end{tikzpicture}
\end{center}
\caption{
The type of directed graphical model under consideration. Solid lines denote the generative model $\pT(\bz)\pT(\bx|\bz)$, dashed lines denote the variational approximation $\qPhi(\bz|\bx)$ to the intractable posterior $\pT(\bz|\bx)$. The variational parameters $\bphi$ are learned jointly with the generative model parameters $\bT$.
}
\end{figure}
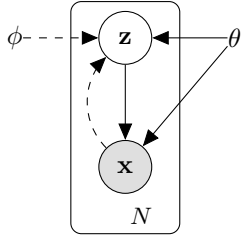

\subsection{Problem scenario}
\label{sec:problem}

Let us consider some dataset $\bX = \{\bx^{(i)}\}_{i=1}^N$ consisting of $N$ i.i.d. samples of some continuous or discrete variable $\bx$. We assume that the data are generated by some random process, involving an unobserved continuous random variable $\bz$. The process consists of two steps: (1) a value $\bzi$ is generated from some prior distribution $p_{\bT^*}(\bz)$; (2) a value $\bxi$ is generated from some conditional distribution $p_{\bT^*}(\bx|\bz)$.  We assume that the prior $p_{\bT^*}(\bz)$ and likelihood $p_{\bT^*}(\bx|\bz)$ come from parametric families of distributions $\pT(\bz)$ and $\pT(\bx|\bz)$, and that their PDFs are differentiable almost everywhere w.r.t. both $\bT$ and $\bz$. Unfortunately, a lot of this process is hidden from our view: the true parameters $\bT^*$ as well as the values of the latent variables $\bzi$ are unknown to us.  

Very importantly, we \emph{do not} make the common simplifying assumptions about the marginal or posterior probabilities. Conversely, we are here interested in a general algorithm that even works efficiently in the case of:
\begin{enumerate}
\item \emph{Intractability}: the case where the integral of the marginal likelihood $\pT(\bx) = \int \pT(\bz) \pT(\bx|\bz) \,d\bz$ is intractable (so we cannot evaluate or differentiate the marginal likelihood), where the true posterior density $\pT(\bz|\bx) = \pT(\bx|\bz)\pT(\bz)/\pT(\bx)$ is intractable (so the EM algorithm cannot be used), and where the required integrals for any reasonable mean-field VB algorithm are also intractable. These intractabilities are quite common and appear in cases of moderately complicated likelihood functions $\pT(\bx|\bz)$, e.g. a neural network with a nonlinear hidden layer.
\item \emph{A large dataset}: we have so much data that batch optimization is too costly; we would like to make parameter updates using small minibatches or even single datapoints. Sampling-based solutions, e.g. Monte Carlo EM, would in general be too slow, since it involves a typically expensive sampling loop per datapoint.
\end{enumerate}

We are interested in, and propose a solution to, three related problems in the above scenario:
\begin{enumerate}
\item Efficient approximate ML or MAP estimation for the parameters $\bT$. The parameters can be of interest themselves, e.g. if we are analyzing some natural process. They also allow us to mimic the hidden random process and generate artificial data that resembles the real data.
\item Efficient approximate posterior inference of the latent variable $\bz$ given an observed value $\bx$ for a choice of parameters $\bT$. This is useful for coding or data representation tasks.
\item Efficient approximate marginal inference of the variable $\bx$. This allows us to perform all kinds of inference tasks where a prior over $\bx$ is required. Common applications in computer vision include image denoising, inpainting and super-resolution.
\end{enumerate}

For the purpose of solving the above problems, let us introduce a recognition model $\qPhi(\bz|\bx)$: an approximation to the intractable true posterior $\pT(\bz|\bx)$. Note that in contrast with the approximate posterior in mean-field variational inference, it is not necessarily factorial and its parameters $\bphi$ are not computed from some closed-form expectation. Instead, we'll introduce a method for learning the recognition model parameters $\bphi$ jointly with the generative model parameters $\bT$.

From a coding theory perspective, the unobserved variables $\bz$ have an interpretation as a latent representation or \emph{code}.  In this paper we will therefore also refer to the recognition model $\qPhi(\bz|\bx)$ as a probabilistic \emph{encoder}, since given a datapoint $\bx$ it produces a distribution (e.g. a Gaussian) over the possible values of the code $\bz$ from which the datapoint $\bx$ could have been generated. In a similar vein we will refer to $\pT(\bx|\bz)$ as a probabilistic \emph{decoder}, since given a code $\bz$ it produces a distribution over the possible corresponding values of $\bx$. 

\subsection{The variational bound}
The marginal likelihood is composed of a sum over the marginal likelihoods of individual datapoints $\log \pT(\bx^{(1)}, \cdots, \bx^{(N)}) = \sum_{i=1}^N \log \pT(\bxi)$, which can each be rewritten as:
\begin{align*}
\log \pT(\bxi) = D_{KL}(\qPhi(\bz|\bxi)||\pT(\bz|\bxi)) + \LB{}{\bxi}
\eqnr\end{align*}
The first RHS term is the KL divergence of the approximate from the true posterior. Since this KL-divergence is non-negative, the second RHS term $\LB{}{\bxi}$ is called the (variational) \emph{lower bound} on the marginal likelihood of datapoint $i$, and can be written as:
\begin{align*}
\log \pT(\bxi) \geq \LB{}{\bxi}
&= \Exp{\qPhi(\bz|\bx)}{- \log \qPhi(\bz|\bx) + \log \pT(\bx,\bz)}
\eqnr\label{eq:lowerbound}\end{align*}
which can also be written as:
\begin{align*}
\LB{}{\bxi} = - D_{KL}(\qPhi(\bz|\bxi) || \pT(\bz)) + \Exp{\qPhi(\bz|\bxi)}{\log \pT(\bxi | \bz)}
\eqnr\label{eq:lowerbound2}\end{align*}
We want to differentiate and optimize the lower bound $\LB{}{\bxi}$ w.r.t. both the variational parameters $\bphi$ and generative parameters $\bT$. However, the gradient of the lower bound w.r.t. $\bphi$ is a bit problematic. The usual (na\"ive) Monte Carlo gradient estimator for this type of problem is:
$\nabla_{\bphi} \Exp{\qPhi(\bz)}{f(\bz)} = \Exp{\qPhi(\bz)}{f(\bz) \nabla_{\qPhi(\bz)} \log \qPhi(\bz) } \simeq \frac{1}{L} \sum_{l=1}^L f(\bz) \nabla_{\qPhi(\bzl)} \log \qPhi(\bzl)$ where $\bzl \sim \qPhi(\bz|\bxi)$. This gradient estimator exhibits exhibits very high variance (see e.g. ~\cite{blei2012variational}) and is impractical for our purposes.

\subsection{The SGVB estimator and AEVB algorithm}
\label{subsec:our_estimator}

In this section we introduce a practical estimator of the lower bound and its derivatives w.r.t. the parameters. We assume an approximate posterior in the form $\qPhi(\bz|\bx)$, but please note that the technique can be applied to the case $\qPhi(\bz)$, i.e. where we do not condition on $\bx$, as well. The fully variational Bayesian method for inferring a posterior over the parameters is given in the appendix.

Under certain mild conditions outlined in section~\ref{subsec:thetrick} for a chosen approximate posterior $\qPhi(\bz|\bx)$ we can reparameterize the random variable $\btz \sim \qPhi(\bz|\bx)$ using a differentiable transformation $\gPhi(\beps,\bx)$ of an (auxiliary) noise variable $\beps$:
\begin{align*}
\btz = \gPhi(\beps,\bx) \text{\quad with \quad} \beps \sim p(\beps)
\eqnr\end{align*}
See section~\ref{subsec:thetrick} for general strategies for chosing such an approriate distribution $p(\beps)$ and function $\gPhi(\beps,\bx)$. 
We can now form Monte Carlo estimates of expectations of some function $f(\bz)$ w.r.t. $\qPhi(\bz|\bx)$ as follows:
\begin{align}
\Exp{\qPhi(\bz|\bxi)}{f(\bz)}
= \Exp{p(\beps)}{f(\gPhi(\beps,\bxi))}
&\simeq \frac{1}{L} \sum_{l=1}^L {f(\gPhi(\bepsl,\bxi))}
 \text{\quad where \quad} \bepsl \sim p(\beps)
\end{align}
We apply this technique to the variational lower bound (eq.~\eqref{eq:lowerbound}), yielding our generic Stochastic Gradient Variational Bayes (SGVB) estimator $\LBT{A}{\bxi} \simeq \LB{}{\bxi}$:
\begin{align*}
\LBT{A}{\bxi}
&= \frac{1}{L} \sum_{l=1}^L 
\log \pT(\bxi, \bzil) - \log \qPhi(\bzil|\bxi) \\
\text{where \quad} \bzil &= \gPhi(\bepsil,\bxi)
\text{\quad and \quad} \bepsl \sim p(\beps)
\eqnr\label{eq:fullestimator}
\end{align*}
Often, the KL-divergence $D_{KL}(\qPhi(\bz|\bxi) || \pT(\bz))$ of eq.~\eqref{eq:lowerbound2} can be integrated analytically (see appendix~\ref{ap:kl_solution}), such that only the expected reconstruction error $\Exp{\qPhi(\bz|\bxi)}{\log \pT(\bxi | \bz)}$ requires estimation by sampling. The KL-divergence term can then be interpreted as regularizing $\bphi$, encouraging the approximate posterior to be close to the prior $\pT(\bz)$.
This yields a second version of the SGVB estimator $\LBT{B}{\bxi} \simeq \LB{}{\bxi}$, corresponding to eq.~\eqref{eq:lowerbound2}, which typically has less variance than the generic estimator:
\begin{align*}
\LBT{B}{\bxi}
&= - D_{KL}(\qPhi(\bz|\bxi) || \pT(\bz))
+ \frac{1}{L} \sum_{l=1}^L (\log \pT(\bxi|\bzil)) \\
\text{where \quad} \bzil &= \gPhi(\bepsil,\bxi)
\text{\quad and \quad} \bepsl \sim p(\beps)
\eqnr\label{eq:estimator2}
\end{align*}
Given multiple datapoints from a dataset $\bX$ with $N$ datapoints, we can construct an estimator of the marginal likelihood lower bound of the full dataset, based on minibatches:
\begin{align*}
\LB{}{\bX} \simeq \LBT{M}{\bX^M} = \frac{N}{M} \sum_{i=1}^M \LBT{}{\bxi}
\eqnr\label{eq:minibatchestimator}\end{align*}
where the minibatch $\bX^M = \{\bxi\}_{i=1}^M$ is a randomly drawn sample of $M$ datapoints from the full dataset $\bX$ with $N$ datapoints. In our experiments we found that the number of samples $L$ per datapoint can be set to $1$ as long as the minibatch size $M$ was large enough, e.g. $M=100$. Derivatives $\nabla_{\bT,\bphi} \widetilde{\mathcal{L}}(\bT;\bX^M)$ can be taken, and the resulting gradients can be used in conjunction with stochastic optimization methods such as SGD or Adagrad~\cite{duchi2010adaptive}. See algorithm~\ref{algorithm} for a basic approach to compute the stochastic gradients.

A connection with auto-encoders becomes clear when looking at the objective function given at eq.~\eqref{eq:estimator2}. The first term is (the KL divergence of the approximate posterior from the prior) acts as a regularizer, while the second term is a an expected negative reconstruction error. 
The function $\gPhi(.)$ is chosen such that it maps a datapoint $\bxi$ and a random noise vector $\bepsl$ to a sample from the approximate posterior for that datapoint: $\bzil = \gPhi(\bepsl, \bxi)$ where $\bzil \sim \qPhi(\bz|\bxi)$. 
Subsequently, the sample $\bzil$ is then input to function $\log \pT(\bxi|\bzil)$, which equals the probability density (or mass) of datapoint $\bxi$ under the generative model, given $\bzil$. This term is a negative \emph{reconstruction error} in auto-encoder parlance.

\begin{algorithm}[t]
\caption{Minibatch version of the Auto-Encoding VB (AEVB) algorithm. Either of the two SGVB estimators in section~\ref{subsec:our_estimator} can be used. We use settings $M=100$ and $L=1$ in experiments. }
\renewcommand{\algorithmicforall}{\textbf{for each}}
\begin{algorithmic}
\State $\bT, \bphi \gets$ Initialize parameters
\Repeat
\State $\bX^M \gets $ Random minibatch of $M$ datapoints (drawn from full dataset)
\State $\beps \gets $ Random samples from noise distribution $p(\beps)$
\State $\bg \gets \nabla_{\bT,\bphi} \LBT{M}{\bX^M,\beps}$ (Gradients of minibatch estimator \eqref{eq:minibatchestimator})
\State $\bT, \bphi \gets $ Update parameters using gradients $\bg$ (e.g. SGD or Adagrad~\cite{duchi2010adaptive})
\Until {convergence of parameters $(\bT,\bphi)$}
\\ \Return $\bT, \bphi$
\end{algorithmic}
\label{algorithm}
\end{algorithm}

\subsection{The reparameterization trick}
\label{subsec:thetrick}

In order to solve our problem we invoked an alternative method for generating samples from $\qPhi(\bz|\bx)$. The essential parameterization trick is quite simple. Let $\bz$ be a continuous random variable, and $\bz \sim \qPhi(\bz|\bx)$ be some conditional distribution. It is then often possible to express the random variable $\bz$ as a deterministic variable $\bz = \gPhi(\beps, \bx)$, where $\beps$ is an auxiliary variable with independent marginal $p(\beps)$, and $\gPhi(.)$ is some vector-valued function parameterized by $\bphi$.

This reparameterization is useful for our case since it can be used to rewrite an expectation w.r.t $\qPhi(\bz|\bx)$ such that the Monte Carlo estimate of the expectation is differentiable w.r.t. $\bphi$. A proof is as follows. Given the deterministic mapping $\bz = \gPhi(\beps, \bx)$ we know that $\qPhi(\bz|\bx) \prod_i d z_i = p(\beps) \prod_i d \epsilon_i$. Therefore\footnote{Note that for infinitesimals we use the notational convention $d\bz = \prod_i d z_i$}, $\int \qPhi(\bz|\bx) f(\bz) \,d\bz = \int p(\beps) f(\bz) \,d\beps = \int p(\beps) f(\gPhi(\beps, \bx)) \,d\beps$. It follows that a differentiable estimator can be constructed: $\int \qPhi(\bz|\bx) f(\bz) \,d\bz \simeq \frac{1}{L} \sum_{l=1}^L f(\gPhi(\bx, \bepsl))$ where $\bepsl \sim p(\beps)$. In section~\ref{subsec:our_estimator} we applied this trick to obtain a differentiable estimator of the variational lower bound.

Take, for example, the univariate Gaussian case: let $z \sim p(z|x) = \mathcal{N}(\mu, \sigma^2)$. In this case, a valid reparameterization is $z = \mu + \sigma \epsilon$, where $\epsilon$ is an auxiliary noise variable $\epsilon \sim \mathcal{N}(0,1)$. Therefore, $\Exp{\mathcal{N}(z; \mu, \sigma^2)}{f(z)} = \Exp{\mathcal{N}(\epsilon; 0, 1)}{f(\mu + \sigma \epsilon)} \simeq \frac{1}{L} \sum_{l=1}^L f(\mu + \sigma \epsilon^{(l)})$ where $\epsilon^{(l)} \sim \mathcal{N}(0,1)$.

For which $\qPhi(\bz|\bx)$ can we choose such a differentiable transformation $\gPhi(.)$ and auxiliary variable $\beps \sim p(\beps)$? Three basic approaches are:
\begin{enumerate}
\item Tractable inverse CDF. In this case, let $\beps \sim \mathcal{U}(\bzero,\bI)$, and let $\gPhi(\beps,\bx)$ be the inverse CDF of $\qPhi(\bz|\bx)$. Examples: Exponential, Cauchy, Logistic, Rayleigh, Pareto, Weibull, Reciprocal, Gompertz, Gumbel and Erlang distributions.

\item Analogous to the Gaussian example, for any "location-scale" family of distributions we can choose the standard distribution (with $\text{location} =0$, $\text{scale} =1$) as the auxiliary variable $\beps$, and let $g(.)=\text{location}+\text{scale} \cdot \beps$. Examples: Laplace, Elliptical, Student's t, Logistic, Uniform, Triangular and Gaussian distributions.

\item Composition: It is often possible to express random variables as different transformations of auxiliary variables. Examples: Log-Normal (exponentiation of normally distributed variable), Gamma (a sum over exponentially distributed variables), Dirichlet (weighted sum of Gamma variates), Beta, Chi-Squared, and F distributions.
\end{enumerate}

When all three approaches fail, good approximations to the inverse CDF exist requiring computations with time complexity comparable to the PDF (see e.g. ~\cite{devroye1986sample} for some methods).

\section{Example: Variational Auto-Encoder}
\label{sec:example}
In this section we'll give an example where we use a neural network for the probabilistic encoder $\qPhi(\bz|\bx)$ (the approximation to the posterior of the generative model $\pT(\bx, \bz)$) and where the parameters $\bphi$ and $\bT$ are optimized jointly with the AEVB algorithm.

Let the prior over the latent variables be the centered isotropic multivariate Gaussian $\pT(\bz) = \mathcal{N}(\bz; \bzero, \bI)$. Note that in this case, the prior lacks parameters. We let $\pT(\bx|\bz)$ be a multivariate Gaussian (in case of real-valued data) or Bernoulli (in case of binary data) whose distribution parameters are computed from $\bz$ with a MLP (a fully-connected neural network with a single hidden layer, see appendix~\ref{ap:encoders_decoders}). Note the true posterior $\pT(\bz|\bx)$ is in this case intractable.
While there is much freedom in the form $\qPhi(\bz|\bx)$, we'll assume the true (but intractable) posterior takes on a approximate Gaussian form with an approximately diagonal covariance. In this case, we can let the variational approximate posterior be a multivariate Gaussian with a diagonal covariance structure\footnote{Note that this is just a (simplifying) choice, and not a limitation of our method.}:
\begin{align*}
\log \qPhi(\bz|\bxi)
&= \log \mathcal{N}(\bz; \bmu^{(i)}, \bsigma^{2 (i)} \bI)
\eqnr\end{align*}
where the mean and s.d. of the approximate posterior, $\bmu^{(i)}$ and $\bsigma^{(i)}$, are outputs of the encoding MLP, i.e. nonlinear functions of datapoint $\bx^{(i)}$ and the variational parameters $\bphi$ (see appendix~\ref{ap:encoders_decoders}).

As explained in section~\ref{subsec:thetrick}, we sample from the posterior $\bzil \sim \qPhi(\bz|\bxi)$ using $\bzil = \gPhi(\bxi, \bepsl) = \bmu^{(i)} + \bsigma^{(i)} \odot \bepsl$ where $\bepsl \sim \mathcal{N}(\bzero,\bI)$. With $\odot$ we signify an element-wise product.
In this model both $\pT(\bz)$ (the prior) and $\qPhi(\bz|\bx)$ are Gaussian; in this case, we can use the estimator of eq.~\eqref{eq:estimator2} where the KL divergence can be computed and differentiated without estimation (see appendix~\ref{ap:kl_solution}). The resulting estimator for this model and datapoint $\bxi$ is:
\begin{align*}
\LB{}{\bxi}
&\simeq 
\frac{1}{2} \sum_{j=1}^J \left(1 + \log ((\sigma_j^{(i)})^2) - (\mu_j^{(i)})^2 - (\sigma_j^{(i)})^2 \right)
+ \frac{1}{L} \sum_{l=1}^L \log \pT(\bxi|\bzil) \\
\text{where\quad}
\bzil &= \bmu^{(i)} + \bsigma^{(i)} \odot \beps^{(l)}
\text{\quad and \quad}
\bepsl \sim \mathcal{N}(0,\bI)
\label{eq:gaussian_estimator}\eqnr\end{align*}
As explained above and in appendix~\ref{ap:encoders_decoders}, the decoding term $\log \pT(\bxi|\bzil)$ is a Bernoulli or Gaussian MLP, depending on the type of data we are modelling.

\section{Related work}
The wake-sleep algorithm~\cite{hinton1995wake} is, to the best of our knowledge, the only other on-line learning method in the literature that is applicable to the same general class of continuous latent variable models. Like our method, the wake-sleep algorithm employs a recognition model that approximates the true posterior. A drawback of the wake-sleep algorithm is that it requires a concurrent optimization of two objective functions, which together do not correspond to optimization of (a bound of) the marginal likelihood.
An advantage of wake-sleep is that it also applies to models with discrete latent variables. Wake-Sleep has the same computational complexity as AEVB per datapoint.

Stochastic variational inference~\cite{hoffman2013stochastic} has recently received increasing interest. Recently, \cite{blei2012variational} introduced a control variate schemes to reduce the high variance of the na\"ive gradient estimator discussed in section~\ref{sec:problem}, and applied to exponential family approximations of the posterior. In~\cite{ranganath2013black} some general methods, i.e. a control variate scheme, were introduced for reducing the variance of the original gradient estimator. In~\cite{salimans2013fixedform}, a similar reparameterization as in this paper was used in an efficient version of a stochastic variational inference algorithm for learning the natural parameters of exponential-family approximating distributions.

The AEVB algorithm exposes a connection between directed probabilistic models (trained with a variational objective) and auto-encoders. A connection between \emph{linear} auto-encoders and a certain class of generative linear-Gaussian models has long been known. In ~\cite{roweis1998algorithms} it was shown that PCA corresponds to the maximum-likelihood (ML) solution of a special case of the linear-Gaussian model with a prior $p(\bz) = \mathcal{N}(0,\bI)$ and a conditional distribution $p(\bx|\bz) = \mathcal{N}(\bx; \bW \bz, \epsilon \bI)$, specifically the case with infinitesimally small $\epsilon$. 

In relevant recent work on autoencoders~\cite{vincent2010stacked} it was shown that the training criterion of unregularized autoencoders corresponds to maximization of a lower bound (see the infomax principle~\cite{linsker1989application}) of the mutual information between input $X$ and latent representation $Z$. Maximizing (w.r.t. parameters) of the mutual information is equivalent to maximizing the conditional entropy, which is lower bounded by the expected loglikelihood of the data under the autoencoding model~\cite{vincent2010stacked}, i.e. the negative reconstrution error.
However, it is well known that this reconstruction criterion is in itself not sufficient for learning useful representations~\cite{bengio2013representation}.
Regularization techniques have been proposed to make autoencoders learn useful representations, such as  denoising, contractive and sparse autoencoder variants~ \cite{bengio2013representation}. The SGVB objective contains a regularization term dictated by the variational bound (e.g. eq.~\eqref{eq:gaussian_estimator}), lacking the usual nuisance regularization hyperparameter required to learn useful representations.
Related are also encoder-decoder architectures such as the predictive sparse decomposition (PSD)~\cite{koray-psd-08}, from which we drew some inspiration. Also relevant are the recently introduced Generative Stochastic Networks~\cite{bengio2013deep} where noisy auto-encoders learn the transition operator of a Markov chain that samples from the data distribution. In~\cite{salakhutdinov2010efficient} a recognition model was employed for efficient learning with Deep Boltzmann Machines.
These methods are targeted at either unnormalized models (i.e. undirected models like Boltzmann machines) or limited to sparse coding models, in contrast to our proposed algorithm for learning a general class of directed probabilistic models.

The recently proposed DARN method ~\cite{gregor2013deep}, also learns a directed probabilistic model using an auto-encoding structure, however their method applies to binary latent variables. 
Even more recently, ~\cite{rezende2014stochastic} also make the connection between auto-encoders, directed proabilistic models and stochastic variational inference using the reparameterization trick we describe in this paper. Their work was developed independently of ours and provides an additional perspective on AEVB.

\begin{figure}[t]
\includegraphics[width=0.95\columnwidth]{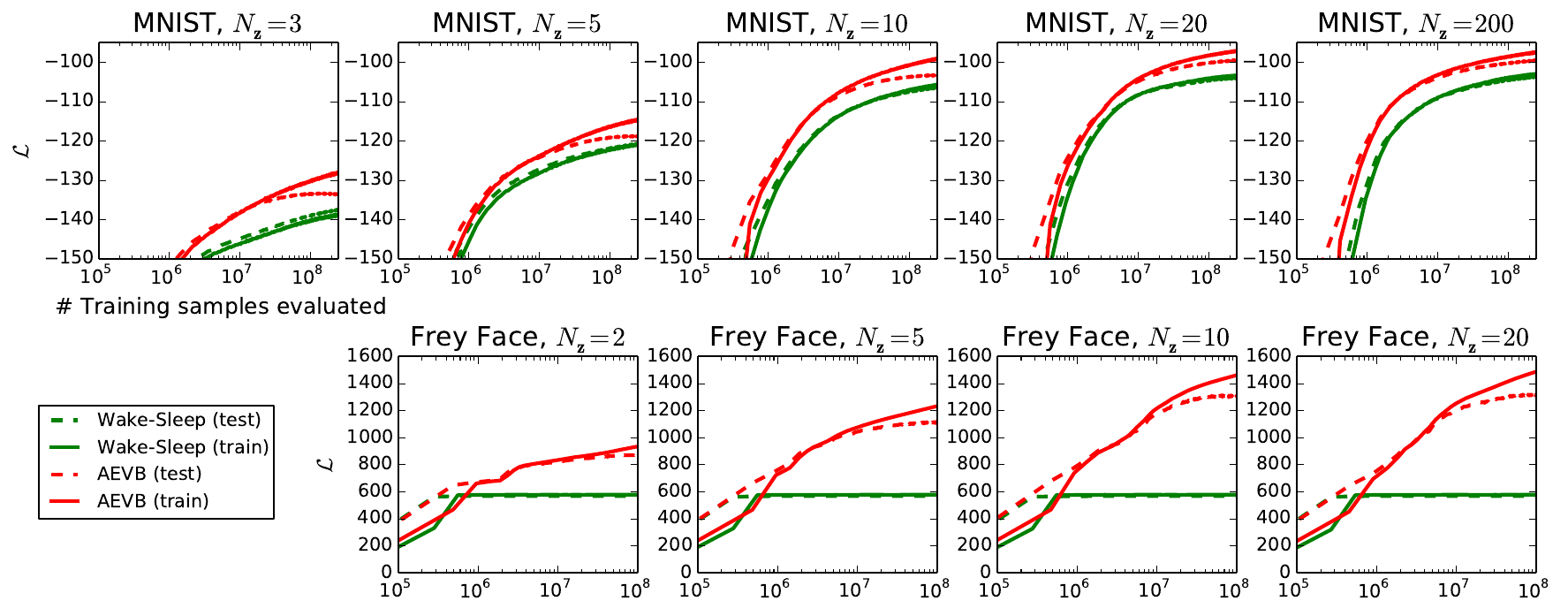}
\caption{Comparison of our AEVB method to the wake-sleep algorithm, in terms of optimizing the lower bound, for different dimensionality of latent space ($N_{\bz}$). Our method converged considerably faster and reached a better solution in all experiments. Interestingly enough, more latent variables does not result in more overfitting, which is explained by the regularizing effect of the lower bound. Vertical axis: the estimated average variational lower bound per datapoint. The estimator variance was small ($< 1$) and omitted. Horizontal axis: amount of training points evaluated. Computation took around 20-40 minutes per million training samples with a Intel Xeon CPU running at an effective 40 GFLOPS.}\label{fig:lowbound}
\end{figure}

\section{Experiments}

We trained generative models of images from the MNIST and Frey Face datasets\footnote{Available at \url{http://www.cs.nyu.edu/~roweis/data.html}} and compared learning algorithms in terms of the variational lower bound, and the estimated marginal likelihood. 

The generative model (encoder) and variational approximation (decoder) from section~\ref{sec:example} were used, where the described encoder and decoder have an equal number of hidden units. Since the Frey Face data are continuous, we used a decoder with Gaussian outputs, identical to the encoder, except that the means were constrained to the interval $(0,1)$ using a sigmoidal activation function at the decoder output. 
Note that with \emph{hidden units} we refer to the hidden layer of the neural networks of the encoder and decoder.

Parameters are updated using stochastic gradient ascent where gradients are computed by differentiating the lower bound estimator $\nabla_{\bT,\bphi} \LB{}{\bX}$ (see algorithm ~\ref{algorithm}), plus a small weight decay term corresponding to a prior $p(\bT) = \mathcal{N}(0,\bI)$.  Optimization of this objective is equivalent to approximate MAP estimation, where the likelihood gradient is approximated by the gradient of the lower bound.

We compared performance of AEVB to the wake-sleep algorithm~\cite{hinton1995wake}.  We employed the same encoder (also called recognition model) for the wake-sleep algorithm and the variational auto-encoder. All parameters, both variational and generative, were initialized by random sampling from $\mathcal{N}(0,0.01)$, and were jointly stochastically optimized using the MAP criterion. Stepsizes were adapted with Adagrad~\cite{duchi2010adaptive}; the Adagrad global stepsize parameters were chosen from \{0.01, 0.02, 0.1\} based on performance on the training set in the first few iterations. Minibatches of size $M=100$ were used, with $L=1$ samples per datapoint.

\paragraph{Likelihood lower bound} 
We trained generative models (decoders) and corresponding encoders (a.k.a. recognition models) having $500$ hidden units in case of MNIST, and $200$ hidden units in case of the Frey Face dataset (to prevent overfitting, since it is a considerably smaller dataset). The chosen number of hidden units is based on prior literature on auto-encoders, and the relative performance of different algorithms was not very sensitive to these choices. Figure~\ref{fig:lowbound} shows the results when comparing the lower bounds. Interestingly, superfluous latent variables did not result in overfitting, which is explained by the regularizing nature of the variational bound.

\begin{figure}[t]
\begin{subfigure}[t]{1\textwidth}
\includegraphics[width=1\columnwidth]{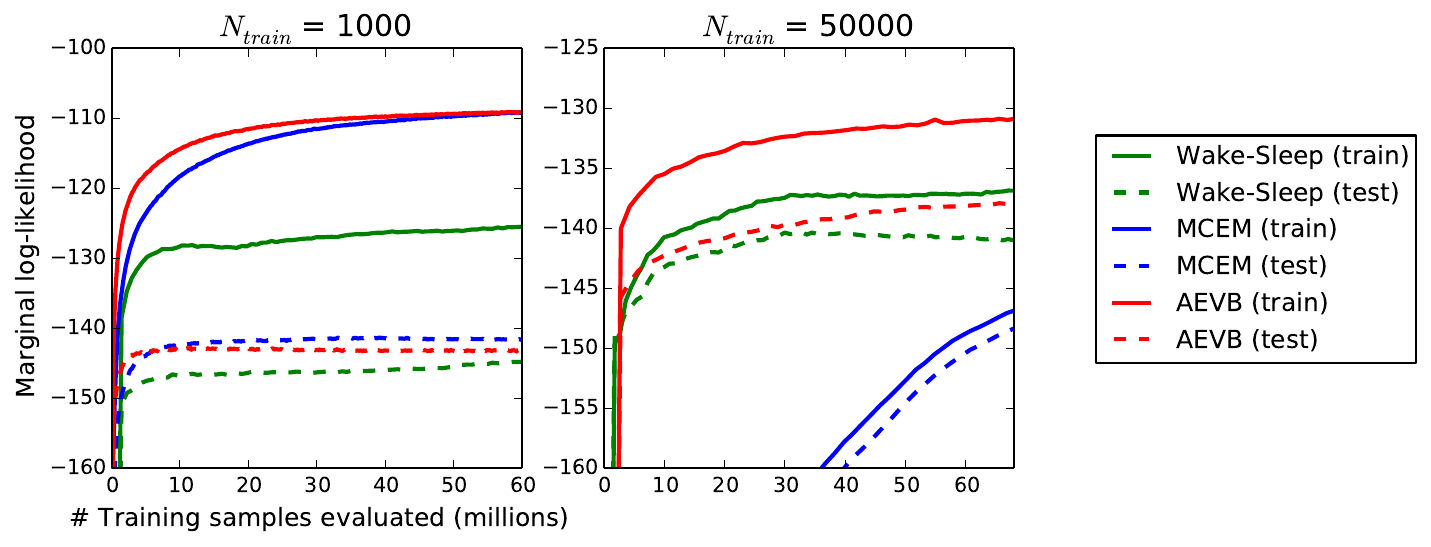}
\end{subfigure}
\caption{Comparison of AEVB to the wake-sleep algorithm and Monte Carlo EM, in terms of the estimated marginal likelihood, for a different number of training points. Monte Carlo EM is not an on-line algorithm, and (unlike AEVB and the wake-sleep method) can't be applied efficiently for the full MNIST dataset.
}\label{fig:marglik}
\end{figure}

\paragraph{Marginal likelihood}
For very low-dimensional latent space it is possible to estimate the marginal likelihood of the learned generative models using an MCMC estimator. More information about the marginal likelihood estimator is available in the appendix. For the encoder and decoder we again used neural networks, this time with 100 hidden units, and 3 latent variables; for higher dimensional latent space the estimates became unreliable. Again, the MNIST dataset was used.
The AEVB and Wake-Sleep methods were compared to Monte Carlo EM (MCEM) with a Hybrid Monte Carlo (HMC)~\cite{duane1987hybrid} sampler; details are in the appendix. We compared the convergence speed for the three algorithms, for a small and large training set size. Results are in figure~\ref{fig:marglik}. 

\paragraph{Visualisation of high-dimensional data} If we choose a low-dimensional latent space (e.g. 2D), we can use the learned encoders (recognition model) to project high-dimensional data to a low-dimensional manifold. See appendix~\ref{ap:visualisations} for visualisations of the 2D latent manifolds for the MNIST and Frey Face datasets.

\section{Conclusion}
We have introduced a novel estimator of the variational lower bound, Stochastic Gradient VB (SGVB), for efficient approximate inference with continuous latent variables. The proposed estimator can be straightforwardly differentiated and optimized using standard stochastic gradient methods. For the case of i.i.d. datasets and continuous latent variables per datapoint we introduce an efficient algorithm for efficient inference and learning, Auto-Encoding VB (AEVB), that learns an approximate inference model using the SGVB estimator. The theoretical advantages are reflected in experimental results. 

\section{Future work}
Since the SGVB estimator and the AEVB algorithm can be applied to almost any inference and learning problem with continuous latent variables, there are plenty of future directions: (i) learning hierarchical generative architectures with deep neural networks (e.g. convolutional networks) used for the encoders and decoders, trained jointly with AEVB; (ii) time-series models (i.e. dynamic Bayesian networks); (iii) application of SGVB to the global parameters; (iv) supervised models with latent variables, useful for learning complicated noise distributions.

\clearpage

\bibliography{bib}

\bibliographystyle{alpha} %

\appendix
\begin{figure}[t]
	\centering
	\begin{subfigure}[t]{0.33\textwidth}
		\includegraphics[width=\textwidth]{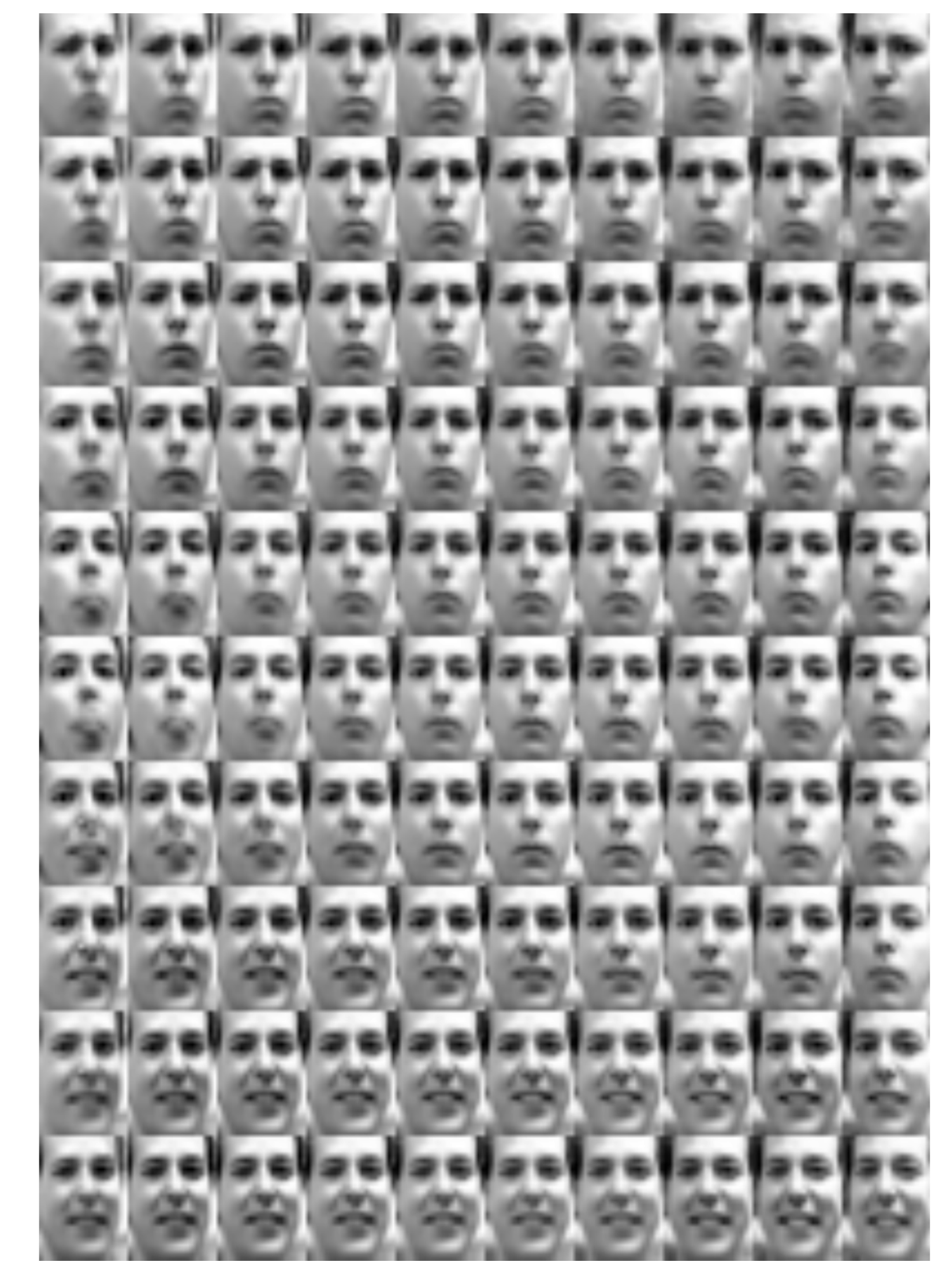}
		\caption{Learned Frey Face manifold}
	\end{subfigure}%
        ~ %
	\begin{subfigure}[t]{0.57\textwidth}
		\includegraphics[width=\textwidth]{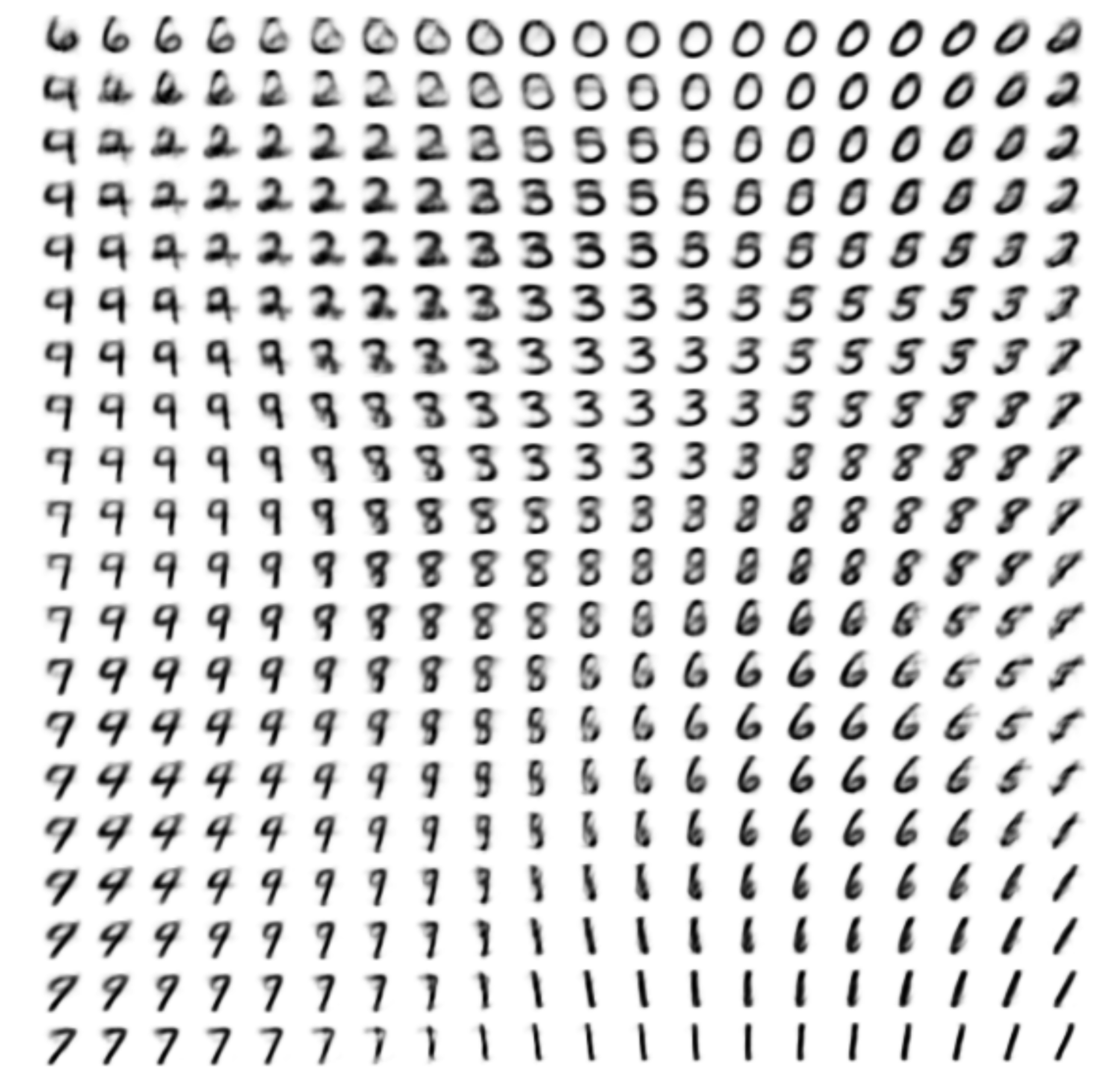}
		\caption{Learned MNIST manifold}
	\end{subfigure}
	\caption{Visualisations of learned data manifold for generative models with two-dimensional latent space, learned with AEVB. Since the prior of the latent space is Gaussian, linearly spaced coordinates on the unit square were transformed through the inverse CDF of the Gaussian to produce values of the latent variables $\bz$. For each of these values $\bz$, we plotted the corresponding generative $\pT(\bx|\bz)$ with the learned parameters $\bT$.}
\label{fig:2dmanifolds}
\end{figure}

\begin{figure}[h]
	\centering
	\begin{subfigure}[t]{0.23\textwidth}
		\includegraphics[width=\textwidth]{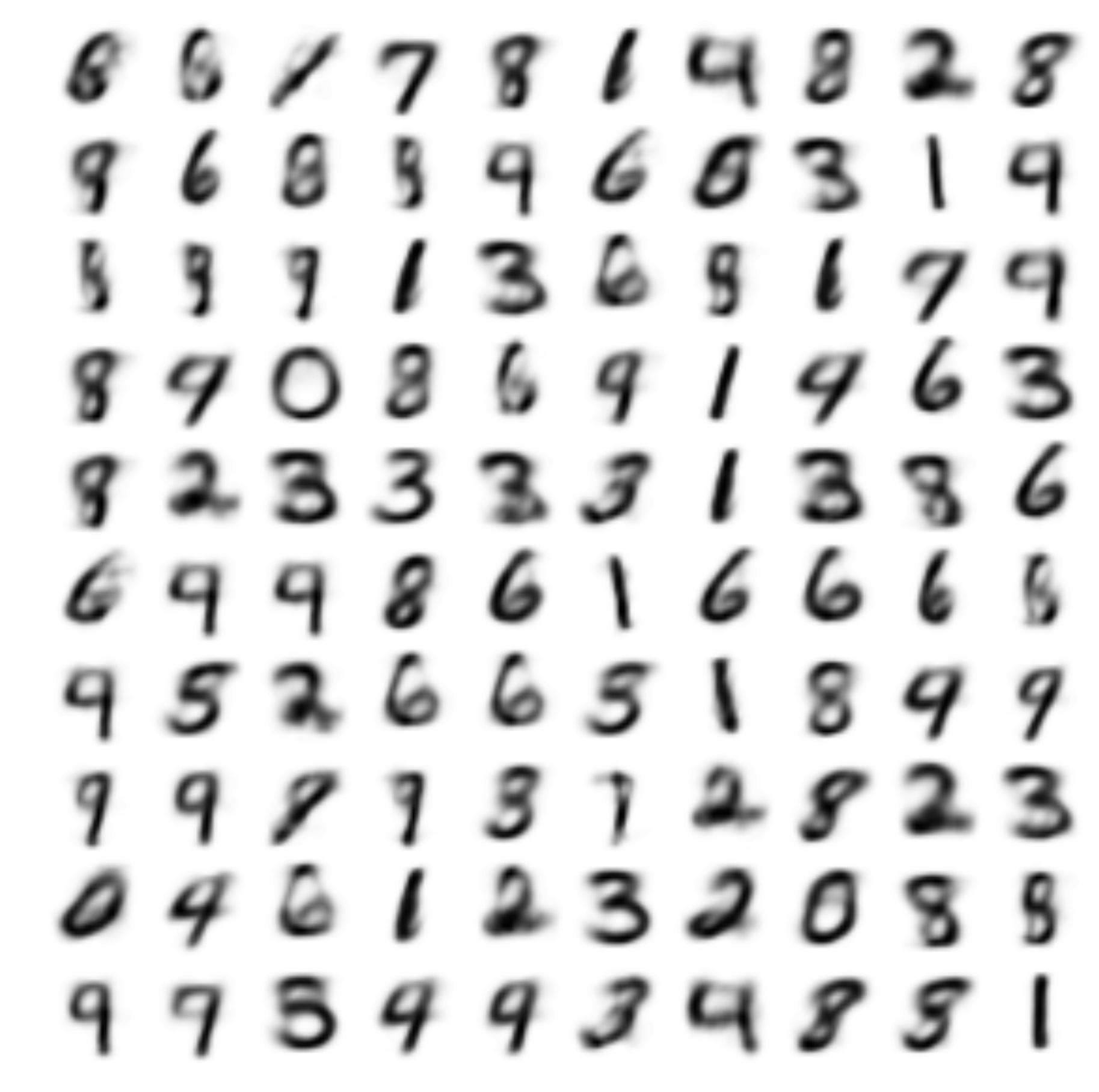}
		\caption{2-D latent space}
	\end{subfigure}
	\begin{subfigure}[t]{0.23\textwidth}
		\includegraphics[width=\textwidth]{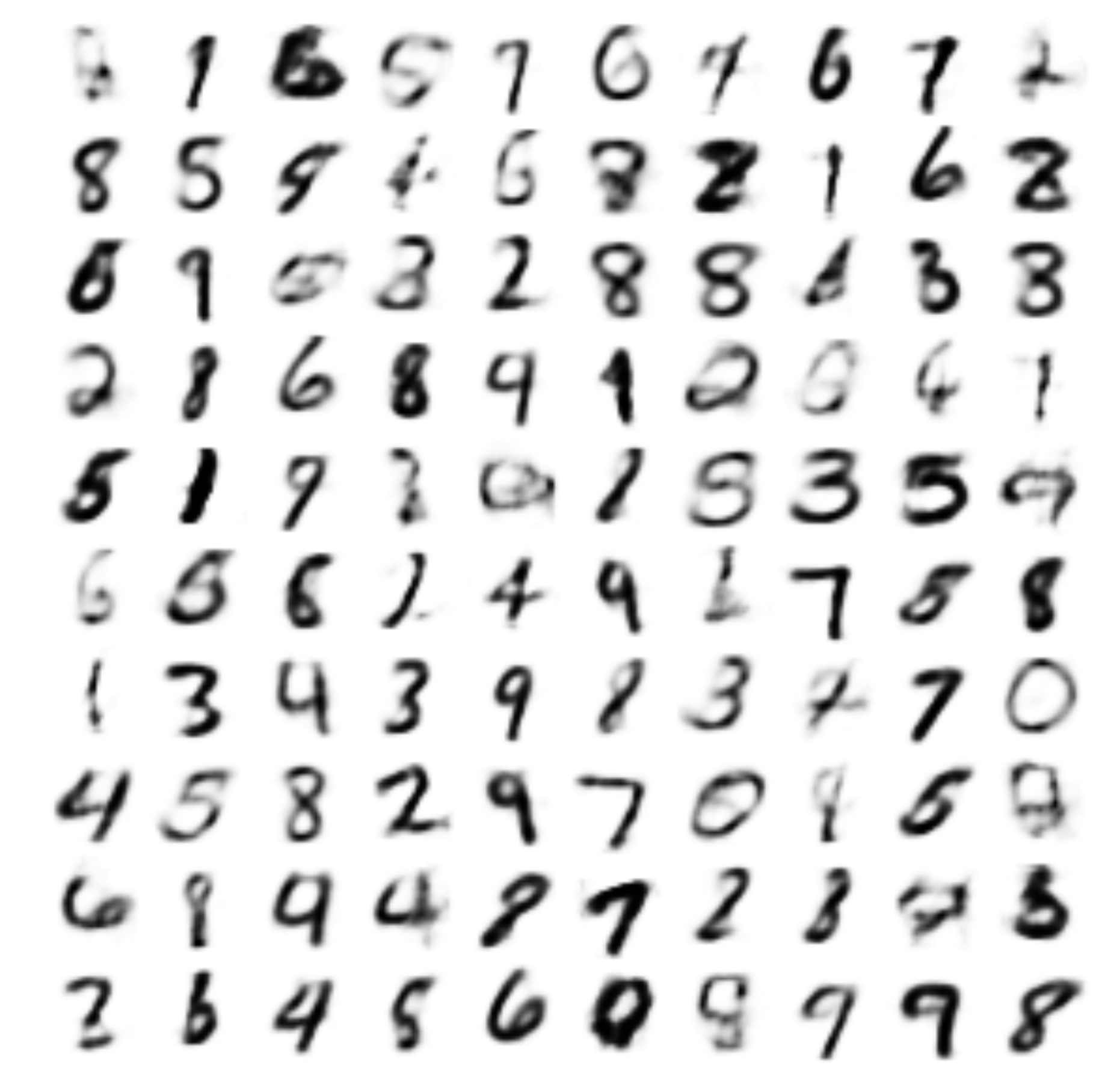}
		\caption{5-D latent space}
	\end{subfigure}
	\begin{subfigure}[t]{0.23\textwidth}
		\includegraphics[width=\textwidth]{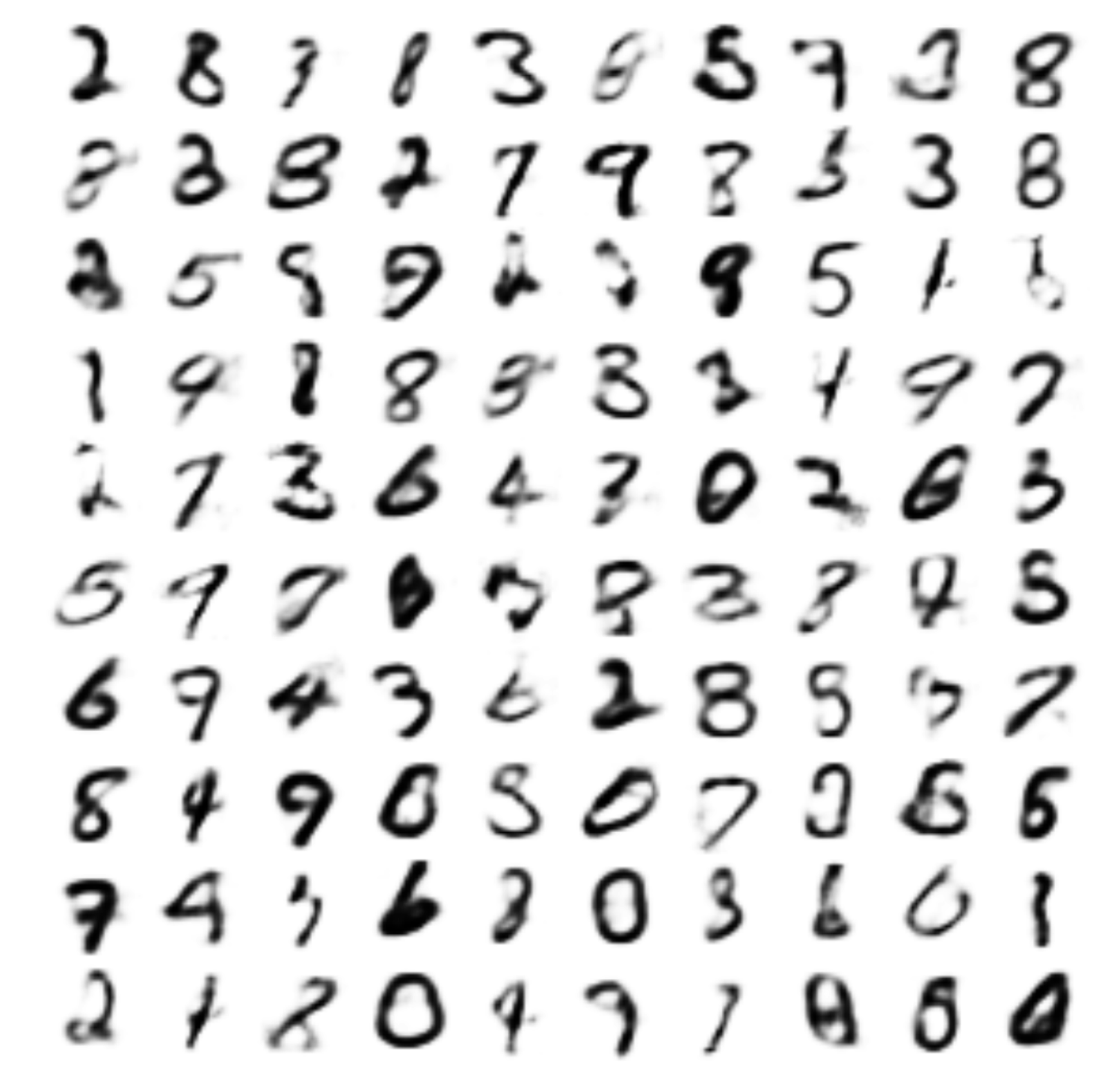}
		\caption{10-D latent space}
	\end{subfigure}
	\begin{subfigure}[t]{0.23\textwidth}
		\includegraphics[width=\textwidth]{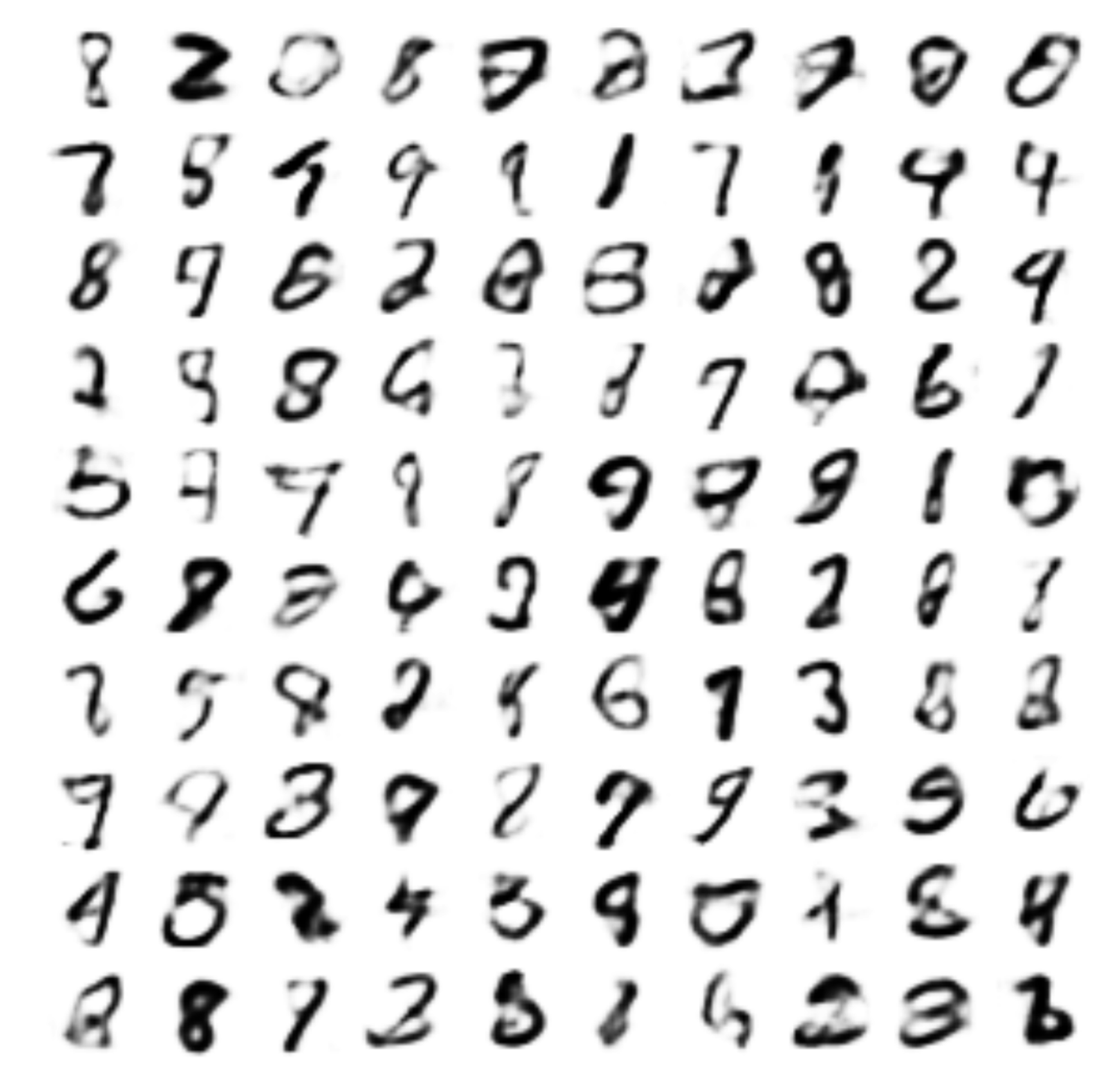}
		\caption{20-D latent space}
	\end{subfigure}
	\caption{Random samples from learned generative models of MNIST for different dimensionalities of latent space.}
\label{fig:mnistsamples}
\end{figure}

\section{Visualisations}
\label{ap:visualisations}
See figures~\ref{fig:2dmanifolds} and~\ref{fig:mnistsamples} for visualisations of latent space and corresponding observed space of models learned with SGVB.

\section{Solution of $- D_{KL}(\qPhi(\bz) || \pT(\bz))$, Gaussian case}
\label{ap:kl_solution}
The variational lower bound (the objective to be maximized) contains a KL term that can often be integrated analytically. Here we give the solution when both the prior $\pT(\bz) = \mathcal{N}(0,\bI)$ and the posterior approximation $\qPhi(\bz|\bxi)$ are Gaussian. Let $J$ be the dimensionality of $\bz$. Let $\bmu$ and $\bsigma$ denote the variational mean and s.d. evaluated at datapoint $i$, and let $\mu_j$ and $\sigma_j$ simply denote the $j$-th element of these vectors. Then:
\begin{align*}
\int \qT(\bz) \log p(\bz) \,d\bz
&= \int \mathcal{N}(\bz;\bmu,\bsigma^2) \log \mathcal{N}(\bz;\bzero,\bI) \,d\bz \\
&=  - \frac{J}{2} \log (2 \pi) - \frac{1}{2} \sum_{j=1}^J (\mu_j^2 + \sigma_j^2)
\end{align*}
And:
\begin{align*}
\int \qT(\bz) \log \qT(\bz) \,d\bz
&= \int \mathcal{N}(\bz;\bmu,\bsigma^2) \log \mathcal{N}(\bz;\bmu,\bsigma^2) \,d\bz \\
&= - \frac{J}{2} \log (2 \pi) - \frac{1}{2} \sum_{j=1}^J ( 1 + \log \sigma^2_j )
\end{align*}
Therefore:
\begin{align*}
- D_{KL}((\qPhi(\bz) || \pT(\bz)) &= \int \qT(\bz) \left(\log \pT(\bz) - \log \qT(\bz)\right) \,d\bz \\
&= \frac{1}{2} \sum_{j=1}^J \left(1 + \log ((\sigma_j)^2) - (\mu_j)^2 - (\sigma_j)^2 \right)
\end{align*}
When using a recognition model $\qPhi(\bz|\bx)$ then $\bmu$ and s.d. $\bsigma$ are simply functions of $\bx$ and the variational parameters $\bphi$, as exemplified in the text.

\section{MLP's as probabilistic encoders and decoders}
\label{ap:encoders_decoders}
In variational auto-encoders, neural networks are used as probabilistic encoders and decoders. There are many possible choices of encoders and decoders, depending on the type of data and model. In our example we used relatively simple neural networks, namely multi-layered perceptrons (MLPs). For the encoder we used a MLP with Gaussian output, while for the decoder we used MLPs with either Gaussian or Bernoulli outputs, depending on the type of data.

\subsection{Bernoulli MLP as decoder}
In this case let $\pT(\bx|\bz)$ be a multivariate Bernoulli whose probabilities are computed from $\bz$ with a fully-connected neural network with a single hidden layer:
\begin{align*}
\log p(\bx|\bz)
&= \sum_{i=1}^D x_i \log y_i + (1-x_i) \cdot \log(1-y_i) \\
\text{where\,\,\,}
\by &= f_\sigma(\bW_2 \tanh(\bW_1 \bz + \bbb_1) + \bbb_2)
\eqnr\end{align*}
where $f_\sigma(.)$ is the elementwise sigmoid activation function, and where $\bT = \{\bW_1, \bW_2, \bbb_1, \bbb_2\}$ are the weights and biases of the MLP.

\subsection{Gaussian MLP as encoder or decoder}
In this case let encoder or decoder be a multivariate Gaussian with a diagonal covariance structure:
\begin{align*}
\log p(\bx|\bz)
&= \log \mathcal{N}(\bx; \bmu, \bsigma^2 \bI)
\\
\text{where\,\,\,}
\bmu &= \bW_4 \bh  +\bbb_4 \\
\log \bsigma^2 &= \bW_5 \bh + \bbb_5 \\
\bh &= \tanh(\bW_3 \bz + \bbb_3)
\eqnr\end{align*}
where $\{\bW_3, \bW_4, \bW_5, \bbb_3, \bbb_4, \bbb_5 \}$ are the weights and biases of the MLP and part of $\bT$ when used as decoder. Note that when this network is used as an encoder $\qPhi(\bz|\bx)$, then $\bz$ and $\bx$ are swapped, and the weights and biases are variational parameters $\bphi$.

\section{Marginal likelihood estimator}
We derived the following marginal likelihood estimator that produces good estimates of the marginal likelihood as long as the dimensionality of the sampled space is low (less then 5 dimensions), and sufficient samples are taken.  Let $\pT(\bx, \bz) = \pT(\bz) \pT(\bx|\bz)$ be the generative model we are sampling from, and for a given datapoint $\bxi$ we would like to estimate the marginal likelihood $\pT(\bxi)$.

The estimation process consists of three stages:
\begin{enumerate}
\item Sample $L$ values $\{\bzl\}$ from the posterior using gradient-based MCMC, e.g. Hybrid Monte Carlo, using $\nabla_{\bz} \log \pT(\bz|\bx) = \nabla_{\bz} \log \pT(\bz) + \nabla_{\bz} \log \pT(\bx|\bz)$.
\item Fit a density estimator $q(\bz)$ to these samples $\{\bzl\}$.
\item Again, sample $L$ new values from the posterior. Plug these samples, as well as the fitted $q(\bz)$, into the following estimator:
\begin{align*}
\pT(\bxi) \simeq \left( \frac{1}{L} \sum_{l=1}^L \frac{q(\bzl)}{\pT(\bz) \pT(\bxi|\bzl)} \right)^{-1} \text{\quad where \quad} \bzl \sim \pT(\bz|\bxi)
\end{align*}
\end{enumerate}

Derivation of the estimator:
\begin{align*}
\frac{1}{\pT(\bxi)} 
&= \frac{\int q(\bz) \,d\bz}{\pT(\bxi)}
= \frac{\int q(\bz) \frac{\pT(\bxi, \bz) }{\pT(\bxi, \bz)} \,d\bz}{\pT(\bxi)} \\
&= \int \frac{\pT(\bxi, \bz)}{\pT(\bxi)} \frac{q(\bz)}{\pT(\bxi, \bz)} \,d\bz \\
&= \int \pT(\bz|\bxi) \frac{q(\bz)}{\pT(\bxi, \bz)} \,d\bz \\
&\simeq \frac{1}{L} \sum_{l=1}^L \frac{q(\bzl)}{\pT(\bz) \pT(\bxi|\bzl)} \text{\quad where \quad} \bzl \sim \pT(\bz|\bxi)
\end{align*}

\section{Monte Carlo EM}
The Monte Carlo EM algorithm does not employ an encoder, instead it samples from the posterior of the latent variables using gradients of the posterior computed with $\nabla_{\bz} \log \pT(\bz|\bx) = \nabla_{\bz} \log \pT(\bz) + \nabla_{\bz} \log \pT(\bx|\bz)$. The Monte Carlo EM procedure consists of 10 HMC leapfrog steps with an automatically tuned stepsize such that the acceptance rate was 90\%, followed by 5 weight updates steps using the acquired sample. For all algorithms the parameters were updated using the Adagrad stepsizes (with accompanying annealing schedule).

The marginal likelihood was estimated with the first 1000 datapoints from the train and test sets, for each datapoint sampling 50 values from the posterior of the latent variables using Hybrid Monte Carlo with 4 leapfrog steps. 

\section{Full VB}
As written in the paper, it is possible to perform variational inference on both the parameters $\bT$ and the latent variables $\bz$, as opposed to just the latent variables as we did in the paper. Here, we'll derive our estimator for that case.

Let $\pA(\bT)$  be some hyperprior for the parameters introduced above, parameterized by $\balpha$. The marginal likelihood can be written as:
\begin{align*}
\log \pA(\bX) = D_{KL}(\qPhi(\bT)||\pA(\bT|\bX)) + \mathcal{L}(\bphi;\bX)
\eqnr\end{align*}
where the first RHS term denotes a KL divergence of the approximate from the true posterior, and where $\mathcal{L}(\bphi;\bX)$ denotes the variational lower bound to the marginal likelihood:
\begin{align*}
\mathcal{L}(\bphi;\bX) = \int \qPhi(\bT) \left( \log \pT(\bX) + \log \pA(\bT) - \log \qPhi(\bT) \right) \,d\bT
\eqnr\label{eq:lowerbound1_ap}\end{align*}
Note that this is a lower bound since the KL divergence is non-negative; the bound equals the true marginal when the approximate and true posteriors match exactly.
The term $\log \pT(\bX)$ is composed of a sum over the marginal likelihoods of individual datapoints $\log \pT(\bX) = \sum_{i=1}^N \log \pT(\bxi)$, which can each be rewritten as:
\begin{align*}
\log \pT(\bxi) = D_{KL}(\qPhi(\bz|\bxi)||\pT(\bz|\bxi)) + \mathcal{L}(\bT,\bphi;\bxi)
\eqnr\end{align*}
where again the first RHS term is the KL divergence of the approximate from the true posterior, and $\mathcal{L}(\bT,\bphi;\bx)$ is the variational lower bound of the marginal likelihood of datapoint $i$:
\begin{align*}
\mathcal{L}(\bT,\bphi;\bxi)
&= \int \qPhi(\bz|\bx) \left(\log \pT(\bxi | \bz) + \log \pT(\bz) - \log \qPhi(\bz|\bx)\right) \,d\bz 
\eqnr\label{eq:lowerbound2_ap}\end{align*}

The expectations on the RHS of eqs ~\eqref{eq:lowerbound1_ap} and \eqref{eq:lowerbound2_ap} can obviously be written as a sum of three separate expectations, of which the second and third component can sometimes be analytically solved, e.g. when both $\pT(\bx)$ and $\qPhi(\bz|\bx)$ are Gaussian. For generality we will here assume that each of these expectations is intractable.

Under certain mild conditions outlined in section (see paper) for chosen approximate posteriors $\qPhi(\bT)$ and $\qPhi(\bz|\bx)$ we can reparameterize conditional samples $\btz \sim \qPhi(\bz|\bx)$ as
\begin{align*}
\btz = \gPhi(\beps,\bx) \text{\quad with \quad} \beps \sim p(\beps)
\eqnr\end{align*}
where we choose a prior $p(\beps)$ and a function $\gPhi(\beps,\bx)$ such that the following holds:
\begin{align*}
\mathcal{L}(\bT,\bphi;\bxi)
&= \int \qPhi(\bz|\bx) \left(\log \pT(\bxi | \bz) + \log \pT(\bz) - \log \qPhi(\bz|\bx)\right) \,d\bz \\
&= \int p(\beps) \left(\log \pT(\bxi | \bz) + \log \pT(\bz) - \log \qPhi(\bz|\bx)\right) \bigg|_{\bz=\gPhi(\beps,\bxi)} \,d\beps
\eqnr\label{eq:reparameterizedbound2_ap}\end{align*}
The same can be done for the approximate posterior $\qPhi(\bT)$:
\begin{align*}
\btT = \hPhi(\bzeta) \text{\quad with \quad} \bzeta \sim p(\bzeta)
\eqnr\end{align*}
where we, similarly as above, choose a prior $p(\bzeta)$ and a function $\hPhi(\bzeta)$ such that the following holds:
\begin{align*}
\mathcal{L}(\bphi;\bX)
&= \int \qPhi(\bT) \left( \log \pT(\bX) + \log \pA(\bT) - \log \qPhi(\bT) \right) \,d\bT \\
&= \int p(\bzeta) \left( \log \pT(\bX) + \log \pA(\bT) - \log \qPhi(\bT) \right) \bigg|_{\bT=\hPhi(\bzeta)} \,d\bzeta
\eqnr\label{eq:reparameterizedbound1_ap}\end{align*}

For notational conciseness we introduce a shorthand notation $\fPhi(\bx, \bz, \bT)$:
\begin{align}
\fPhi(\bx, \bz, \bT) = N \cdot (\log \pT(\bx | \bz) + \log \pT(\bz) - \log \qPhi(\bz|\bx)) + \log \pA(\bT) - \log \qPhi(\bT)
\label{eq:f_ap}
\end{align}
Using equations ~\eqref{eq:reparameterizedbound1_ap} and ~\eqref{eq:reparameterizedbound2_ap}, the Monte Carlo estimate of the variational lower bound, given datapoint $\bxi$, is:
\begin{align*}
\mathcal{L}(\bphi;\bX)
&\simeq \frac{1}{L} \sum_{l=1}^L \fPhi(\bxl, \gPhi(\bepsl,\bxl), \hPhi(\bzetal))
\eqnr\label{eq:fullestimator_ap}\end{align*}
where $\bepsl \sim p(\beps)$ and $\bzetal \sim p(\bzeta)$. 
The estimator only depends on samples from $p(\beps)$ and $p(\bzeta)$ which are obviously not influenced by $\bphi$, therefore the estimator can be differentiated w.r.t. $\bphi$. The resulting stochastic gradients can be used in conjunction with stochastic optimization methods such as SGD or Adagrad~\cite{duchi2010adaptive}. See algorithm~\ref{algorithm} for a basic approach to computing stochastic gradients.

\begin{algorithm}[t]
\caption{Pseudocode for computing a stochastic gradient using our estimator. See text for meaning of the functions $\fPhi$, $\gPhi$ and $\hPhi$. }
\renewcommand{\algorithmicforall}{\textbf{for each}}
\begin{algorithmic}
\Require $\bphi$ (Current value of variational parameters)
\State $\bg \gets 0$
\For{$l$ is $1$ to $L$}
\State $\bx \gets $ Random draw from dataset $\bX$
\State $\beps \gets $ Random draw from prior $p(\beps)$
\State $\bzeta \gets $ Random draw from prior $p(\bzeta)$
\State $\bg \gets \bg + \frac{1}{L} \nabla_{\bphi} \fPhi(\bx, \gPhi(\beps, \bx), \hPhi(\bzeta))$
\EndFor \\
\Return $\bg$
\end{algorithmic}
\label{algorithm_ap}
\end{algorithm}

\subsection{Example}
Let the prior over the parameters and latent variables be the centered isotropic Gaussian $\pA(\bT) = \mathcal{N}(\bz; \bzero, \bI)$ and $\pT(\bz) = \mathcal{N}(\bz; \bzero, \bI)$. Note that in this case, the prior lacks parameters. Let's also assume that the true posteriors are approximatily Gaussian with an approximately diagonal covariance. In this case, we can let the variational approximate posteriors be multivariate Gaussians with a diagonal covariance structure:
\begin{align*}
\log \qPhi(\bT) &= \log \mathcal{N}(\bT; \bmu_{\bT}, \bsigma_{\bT}^2 \bI) \\
\log \qPhi(\bz|\bx)
&= \log \mathcal{N}(\bz; \bmu_{\bz}, \bsigma_{\bz}^2 \bI)
\eqnr\end{align*}
where $\bmu_{\bz}$ and $\bsigma_{\bz}$ are yet unspecified functions of $\bx$.
Since they are Gaussian, we can parameterize the variational approximate posteriors:
\begin{align*}
\qPhi(\bT) &\text{\quad as \quad} \btT = \bmu_{\bT} + \bsigma_{\bT} \odot \bzeta
&\text{\quad where \quad} \bzeta \sim \mathcal{N}(\bzero,\bI) \\
\qPhi(\bz|\bx) &\text{\quad as \quad} \btz = \bmu_{\bz} + \bsigma_{\bz} \odot \beps
&\text{\quad where \quad} \beps \sim \mathcal{N}(\bzero,\bI) 
\end{align*}
With $\odot$ we signify an element-wise product. These can be plugged into the lower bound defined above (eqs \eqref{eq:f_ap} and \eqref{eq:fullestimator_ap}).

In this case it is possible to construct an alternative estimator with a lower variance, since in this model $\pA(\bT)$,  $\pT(\bz)$, $\qPhi(\bT)$ and $\qPhi(\bz|\bx)$ are Gaussian, and therefore four terms of $\fPhi$ can be solved analytically. The resulting estimator is:
\begin{align*}
\mathcal{L}(\bphi;\bX)
&\simeq 
\frac{1}{L} \sum_{l=1}^L 
N \cdot \left( \frac{1}{2} \sum_{j=1}^J \left(1 + \log ((\sigma_{\bz,j}^{(l)})^2) - (\mu_{\bz,j}^{(l)})^2 - (\sigma_{\bz,j}^{(l)})^2 \right)
+ \log \pT(\bxi\bzi) \right) \\
&+ \frac{1}{2} \sum_{j=1}^J \left(1 + \log ((\sigma_{\bT,j}^{(l)})^2) - (\mu_{\bT,j}^{(l)})^2 - (\sigma_{\bT,j}^{(l)})^2 \right)
\label{eq:gaussian_estimator_ap}\eqnr\end{align*}
$\mu_j^{(i)}$ and $\sigma_j^{(i)}$ simply denote the $j$-th element of vectors $\bmu^{(i)}$ and $\bsigma^{(i)}$.

\end{document}